%% file: arxiv.tex
 \documentclass[sigconf,nonacm]{acmart}
\AtBeginDocument{%
  }

\setcopyright{acmlicensed}
\usepackage{graphicx} % DO NOT CHANGE THIS
\usepackage{natbib}  % DO NOT CHANGE THIS AND DO NOT ADD ANY OPTIONS TO IT

\usepackage{algorithm}
\usepackage{amsmath,amsfonts}
\usepackage{algpseudocode}
\usepackage{array}
\usepackage{textcomp}
\usepackage{multicol,multirow,xcolor}
\usepackage{colortbl}

\usepackage{newfloat}
\usepackage{listings}
\begin{document}

\title{From Local Recovery to Unified Generation: Discovering Triple Set in Knowledge Graphs via Observation-Anchored Latent Diffusion}

\author{ \begin{tabular}{c} Jiaqi Wang\textsuperscript{1,2} \quad Yisen Gao\textsuperscript{3} \quad Huayang Shi\textsuperscript{1} \quad Junhao Hou\textsuperscript{1} \quad Wengen Li\textsuperscript{1,*} \\[2pt] Jihong Guan\textsuperscript{1,*} \quad Yichao Zhang\textsuperscript{1} \quad Shuigeng Zhou\textsuperscript{4} \quad Yangqiu Song\textsuperscript{3} \quad Hong Cheng\textsuperscript{2,*} \\[4pt] \small \textsuperscript{1}Tongji University \quad \textsuperscript{2}The Chinese University of Hong Kong \quad \textsuperscript{3}The Hong Kong University of Science and Technology \quad \textsuperscript{4}Fudan University \\[-1pt] \footnotesize \textsuperscript{*}Corresponding authors: \texttt{lwengen@tongji.edu.cn}, \texttt{jhguan@tongji.edu.cn}, \texttt{hcheng@se.cuhk.edu.hk} \end{tabular} }

\renewcommand{\shortauthors}{Wang et al.}
\renewcommand{\shorttitle}{LatentDiff}
\newcommand \modelname{LatentDiff}

\begin{abstract}
Triple set prediction (TSP) is to recover missing relational triples in a knowledge graph by predicting the triple set instead of isolated links. Existing approaches for TSP predominantly rely on the subgraph-based completion paradigm, where the original graph is first partitioned into subgraphs which are then independently completed. However, such a paradigm suffers from fragmented structural reasoning, limited cross subgraph dependency modeling, and missing predictions across subgraphs. This paper proposes \modelname, a novel unified latent generative framework for TSP, which reformulates knowledge graph completion as latent space graph generation. \modelname~ first learns a compact latent representation using a relation-aware residual-quantized knowledge graph autoencoder, which encodes the whole graph into relation-aware latent embeddings, quantizes them with chunked residual codebooks, and reconstructs triples through a factorized decoder. On top of this latent space, we introduce an observation-anchored latent graph diffusion model, where the target latent graph representation is perturbed toward the observed latent, and the reverse process recovers missing triples through progressive latent denoising. Experiments on three benchmark datasets show that \modelname~ outperforms the SOTA in terms of $F_{\rm TSP}$ by 5.4 percentage points on average. Code: https://anonymous.4open.science/r/LKGDiff-44A5. 
\end{abstract}

\begin{CCSXML}
<ccs2012>
   <concept>
       <concept_id>10010147.10010178.10010187</concept_id>
       <concept_desc>Computing methodologies~Knowledge representation and reasoning</concept_desc>
       <concept_significance>500</concept_significance>
       </concept>
   <concept>
       <concept_id>10010147.10010178.10010187.10010188</concept_id>
       <concept_desc>Computing methodologies~Semantic networks</concept_desc>
       <concept_significance>500</concept_significance>
       </concept>
   <concept>
       <concept_id>10010147.10010178.10010187.10010198</concept_id>
       <concept_desc>Computing methodologies~Reasoning about belief and knowledge</concept_desc>
       <concept_significance>100</concept_significance>
       </concept>
 </ccs2012>
\end{CCSXML}

\ccsdesc[500]{Computing methodologies~Knowledge representation and reasoning}
\ccsdesc[500]{Computing methodologies~Semantic networks}
\ccsdesc[100]{Computing methodologies~Reasoning about belief and knowledge}

\keywords{Knowledge Graph, Latent Diffusion Model, Triple Set Prediction}


\maketitle
\section{Introduction}
Knowledge Graphs (KGs) have become fundamental infrastructures for a wide range of intelligent applications~\cite{panllmkgreview,tsang2026autograph,gao2026hypoagent}, including recommendation systems~\cite{jiang2024diffkg}, question answering~\cite{gong2026temp,liu2026cog}, information retrieval~\cite{guo2024lightrag,liang2025kag}, and scientific discovery~\cite{fang2023knowledgenmi,ding2025scitoolagent,yang2026okg}. Despite their broad applicability, real-world KGs are inherently incomplete, motivating extensive research on knowledge graph completion (KGC)~\cite{du2026graphoracle,wan2026tgca}. Traditional KGC methods mainly focus on predicting missing links independently through embedding-based scoring functions~\cite{bordes2013translatintranse,lin2015learningtransr,yang2014dismult,balavzevic2019tucker} or graph neural reasoning mechanisms~\cite{zhang2022redgnn,cui2024promptkgicl}. 
However, in many scenarios, the goal is not to predict a single missing triple, but to recover a set of missing triples that complete the observed graph~\cite{zhang2024gpht}. Thus, Triple Set Prediction (TSP) has recently emerged as a promising direction that achieves KG completion by predicting an entire set of missing triples conditioned on the observed graph~\cite{guan2026difftsp,yuan2024llmtsp}. Compared with conventional link prediction~\cite{cheng2026generating,chen2025diffllm}, TSP emphasizes structured generation and holistic graph completion. TSP can also be formulated as conditional graph generation, a perspective explored by DiffTSP~\cite{guan2026difftsp}.

\begin{figure}
    \centering
    \includegraphics[width=0.9\linewidth]{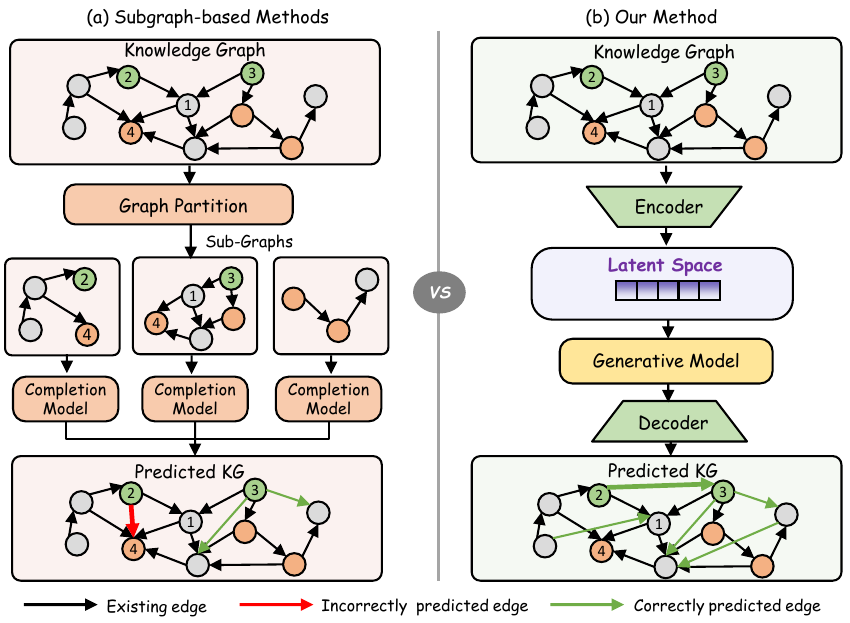}
    \caption{Comparison between subgraph-based completion paradigm and our unified latent generation paradigm. }
    \label{fig:motivation}
\end{figure}

Existing TSP methods predominantly follow a \emph{subgraph-based completion paradigm}. As illustrated in Figure~\ref{fig:motivation}(a), these approaches first partition the original knowledge graph into local subgraphs and then perform completion in each subgraph~\cite{zhang2024gpht,guan2026difftsp}. While such a paradigm reduces computational complexity and improves local reasoning capability, this computational simplification can limit candidate coverage and access to cross-subgraph context.
First, as shown in Figure~\ref{fig:motivation}, it breaks cross-subgraph dependencies, so that a target relation such as the connection between nodes 2 and 3 cannot be recovered by any isolated local completion model. Second, partitioning may discard boundary context. For example, removing the edge between nodes 2 and 1 makes node 2 locally ambiguous and leads to a spurious prediction between nodes 2 and 4. More broadly, independently completed subgraphs lack a unified global latent space, which weakens graph-level semantic coordination and can yield missing or inconsistent predictions after merging.

These limitations motivate a central question: can we recover missing triples across the whole knowledge graph without restricting prediction to subgraphs, while keeping generation computationally manageable? Directly generating a dense multi-relational adjacency remains costly, as its size grows quadratically with the number of entities and linearly with the number of relation types. Our key idea is to compress the state on which generation operates, rather than restrict the prediction domain through graph partitioning.
To this end, we propose \textbf{unified latent generation}. As illustrated in Figure~\ref{fig:motivation}(b), this paradigm performs graph completion in a compact latent space, allowing triples to be generated without requiring their endpoints to co-occur in a local subgraph.

While unified latent modeling avoids the structural fragmentation caused by graph partitioning, recovering the target KG still requires transforming the observed representation into the target one. To instantiate this paradigm, we use conditional latent diffusion to learn recovery from perturbed target representations in TSP. Diffusion performs generation through iterative refinement~\cite{austin2021d3pm,luo2023refusion,luo2023imageirsde}, which is suitable for TSP because missing triples are structurally correlated. The denoising process progressively refines the latent graph representation and adjusts correlated relational patterns among triples~\cite{sheng2026conditional,yisen2026toward}.
However, there are two challenges in applying a generic latent diffusion model to TSP. 
First, a straightforward VQ-VAE~\cite{van2017neuralvqvae} is not sufficient for TSP since KGs are sparse and multi-relational~\cite{ma2025s2dn,ding2025towards,yuan2025knowledge}, while standard VQ-VAE is not designed to preserve relational adjacency patterns. 
Second, standard diffusion~\cite{ho2020ddpm} is not ideal for TSP since the observed KG already provides a strong structural anchor. Standard diffusion gradually perturbs the latent representation toward an isotropic Gaussian prior, which may unnecessarily destroy the structural information already contained in the observed graph.

To address these challenges, we propose \modelname{}, a unified conditional latent generative model for TSP.
For the first challenge, we propose a relation-aware residual-quantized knowledge graph autoencoder. Its encoder constructs latent embeddings from relational adjacency and uses relational view attention to capture dependencies among different relational contexts. To improve the expressiveness of latent space, we use chunked residual vector quantization, which decomposes each latent vector into multiple chunks and quantizes them through multi-level residual vector quantization progressively to enhance reconstruction capacity. Finally, a factorized triple decoder injects relational reasoning bias into KG reconstruction.

For the second challenge, we develop an observation-anchored latent graph diffusion module. Building upon the mean-reverting diffusion principle, its forward process gradually perturbs the target latent representation toward the observed latent. The reverse process is parameterized by a dual-view latent graph denoiser, which captures dependencies along both the relation and entity axes and progressively reconstructs the missing relational information.
Our main contributions include:
    (1) We formulate knowledge graph triple set prediction as a \textbf{unified conditional latent generation} problem, avoiding the candidate space and contextual restrictions induced by subgraph partitioning.
    (2) We propose a \textbf{relation-aware residual-quantized knowledge graph autoencoder} to construct a compact shared latent space for reconstructing relation-specific adjacency patterns.
    (3) We introduce an \textbf{observation-anchored latent graph diffusion model} that combines a mean-reverting diffusion process with a dual-view latent graph denoiser to progressively recover missing relational information in the latent space.  
    (4) Extensive experiments demonstrate that our method outperforms state-of-the-art baselines across multiple datasets.

\section{Related Work}

% \subsection{Knowledge Graph Completion}
% Traditional knowledge graph completion methods mainly rely on embedding-based scoring functions, where entities and relations are embedded into continuous vector spaces and missing triples are predicted via predefined score functions. Representative approaches include TransE~\cite{bordes2013translatintranse},  DistMult~\cite{yang2014dismult}, ComplEx~\cite{trouillon_complex_2016}, HAKE~\cite{zhang2020hake}, PairRE~\cite{chao2020pairre} and TuckER~\cite{balavzevic2019tucker}. More recent methods incorporate graph structure, including RGCN~\cite{2018rgcn}, CompGCN~\cite{CompGCN}, NBFNet~\cite{zhu2021nbfnet}, RED-GNN~\cite{zhang2022redgnn}, AstarNet~\cite{zhu2023astarnet}, KG-ICL~\cite{cui2024promptkgicl}, and ULTRA~\cite{2024ultra}.  These methods have achieved strong performance in link prediction, but mainly focus on individual triples and struggle to capture complex structural dependencies among predicted triples.
\subsection{Triple Set Prediction}
Triple set prediction extends conventional KGC from predicting individual triples to recovering a set of missing triples from an observed KG and provides a more comprehensive setting for automatic KGC. Recent studies address this task from various angles: GPHT~\cite{zhang2024gpht} introduces dedicated evaluation protocols and metrics for this task, LLMTSP~\cite{yuan2024llmtsp} and SMuRT~\cite{yuan2026smurttsp} leverage an LLM to generate the triple set, and DiffTSP~\cite{guan2026difftsp} designs a discrete diffusion model for TSP. To make set-level prediction tractable on KGs, existing models often decompose the original graph into smaller subgraphs and perform local subgraph prediction. However, subgraph partitioning suffers from separating target entities and fragmenting the relational evidence required to predict the connecting triples.

\subsection{Diffusion Models for Graphs}
Diffusion models have demonstrated remarkable success in continuous domains such as image synthesis~\cite{rombach2022ldm,lugmayr2022repaint,xia2023diffir,liu2025effective}  and sequential generation~\cite{nie2025llada}. Several recent works ~\cite{unigraphlatent,boget2025lag1,vignac2022digress,jo2022gdoss} attempt to extend diffusion models to graphs by performing iterative denoising over graph structures or learned representations~\cite{vignac2022digress,jo2022gdoss}. 
Recent studies further perform diffusion in compact latent spaces. LGD~\cite{unigraphlatent} develops an encoder decoder latent diffusion framework for graph generation and prediction, while HypDiff~\cite{yisenhyperbolic} incorporates geometry-aware latent diffusion to better preserve graph topology. These studies mostly focus on generic graph generation or prediction. Triple set prediction instead requires recovering missing relational facts from an observed multi-relational graph, where the latent space must preserve relation-specific structures while supporting conditional graph completion. 
Different from these studies, we focus on conditional triple set recovery in multi-relational KGs.
Beyond graph generation, diffusion models have also been explored for knowledge graph completion. For example, some methods use diffusion to learn triple embeddings for KGC~\cite{kglpdiffwww,long2024kgdm}, while others use diffusion as the hard negative samples generator~\cite{li2026kgc}. 
However, these methods primarily evaluate link prediction, whereas TSP requires selecting a set of missing triples from the candidate space defined by the observed KG.

\section{Problem Definition}
Formally, a knowledge graph is usually represented as $\mathcal{G} = (\mathcal{E}, \mathcal{R}, \mathcal{T})$, where $\mathcal{E}$ and $\mathcal{R}$ denote the entity set and relation set, respectively, and $\mathcal{T} \subseteq \mathcal{E}\times \mathcal{R}\times \mathcal{E}$ is the set of factual triples. Each triple $(h,r,t)\in\mathcal{T}$ describes a relation $r$ from a head entity $h$ to a tail entity $t$.
In TSP, we are given an observed incomplete knowledge graph $ \mathcal{G}_{\mathrm{obs}}= (\mathcal{E}, \mathcal{R}, \mathcal{T}_{\mathrm{obs}})$, where $\mathcal{T}_{\mathrm{obs}}$ is a subset of the triples in the target graph. 
The goal is to recover a set of missing triples $\mathcal{T}_{\mathrm{miss}} =    \mathcal{T}_{\mathrm{tar}}    \setminus \mathcal{T}_{\mathrm{obs}},$ where $    \mathcal{G}_{\mathrm{tar}} =    (\mathcal{E}, \mathcal{R}, \mathcal{T}_{\mathrm{tar}})$ denotes the target KG. 
The predicted KG can then be written as
\begin{equation}
    \mathcal{G}_{\mathrm{pred}}
    =
    (\mathcal{E}, \mathcal{R}, 
    \mathcal{T}_{\mathrm{obs}}\cup \widehat{\mathcal{T}}_{\mathrm{miss}}),
\end{equation}
where $\widehat{\mathcal{T}}_{\mathrm{miss}}$ is the predicted missing triple set.
The learning objective is to generate a triple set $\widehat{\mathcal{T}}_{\mathrm{miss}}$ that approximates the ground-truth missing set $\mathcal{T}_{\mathrm{miss}}$. 
During inference, the predicted triples that already appear in $\mathcal{T}_{\mathrm{obs}}$ are filtered out, and the final output is evaluated against the missing set.

\section{Method}
\begin{figure*}
    \centering
    \includegraphics[width=1\linewidth]{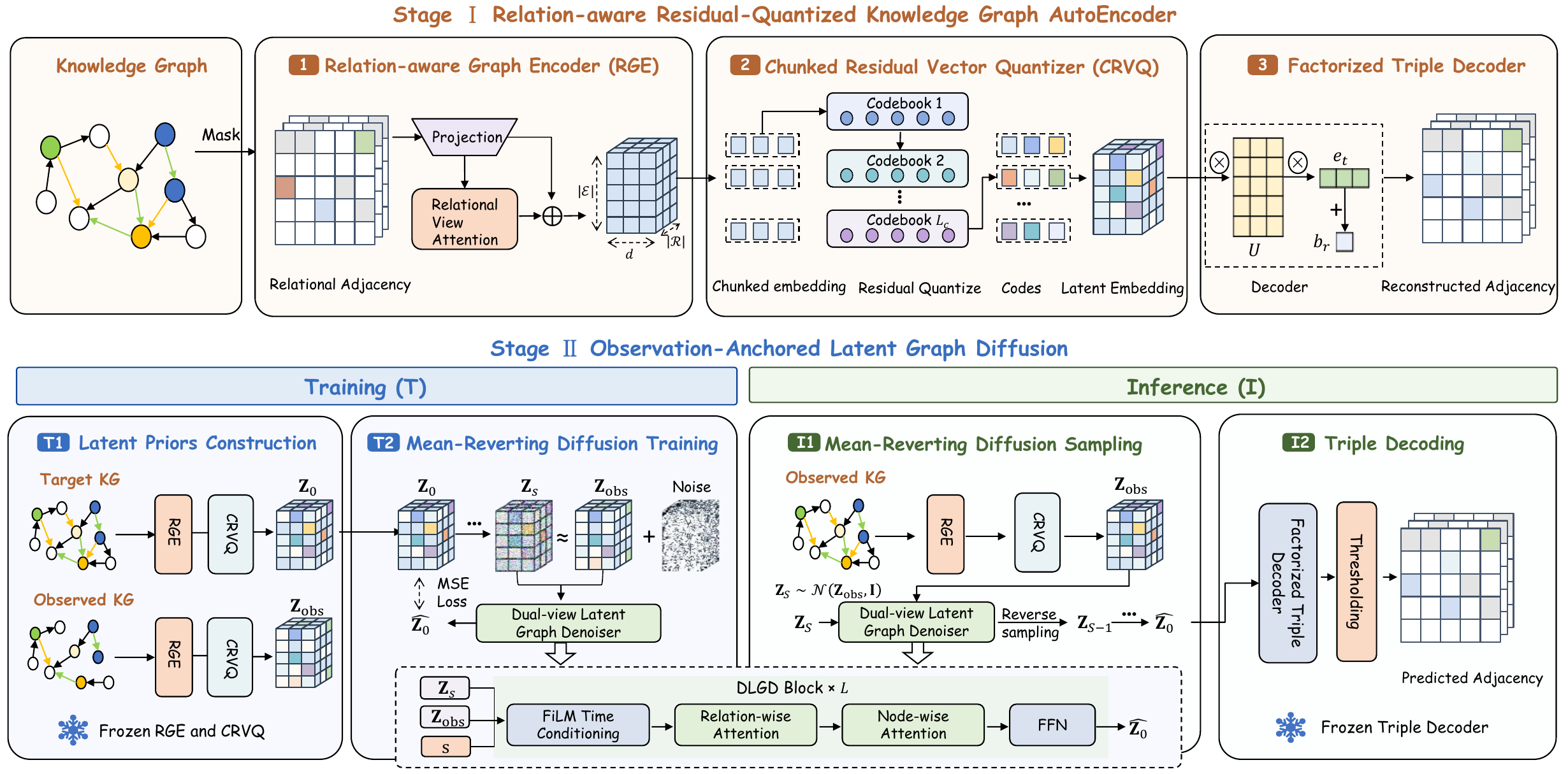}
    \caption{Overview of \modelname{}. Stage I learns a relation-aware residual-quantized latent space for KG reconstruction. Stage II performs observation-anchored latent diffusion to recover the target KG from the observed KG. }
    % \vspace{-5mm}
    \label{fig:model}
\end{figure*}

\subsection{Overview}
As illustrated in Figure~\ref{fig:model}, \modelname{} is a two-stage framework that first learns a compact KG latent space and then performs conditional generation within it. In Stage I, a \textbf{relation-aware residual-quantized knowledge graph autoencoder} (details in Sec.~\ref{sec:KG-autoencoder}), consisting of a relation-aware graph encoder, a chunked residual vector quantizer, and a factorized triple decoder, maps a KG to $\mathbf{Z}\in\mathbb{R}^{|\mathcal{E}|\times|\mathcal{R}|\times d}$, where each embedding $\mathbf{z}_{h,r}$ represents a head-relation context for reconstructing its tail entities. In Stage II, an \textbf{observation-anchored latent graph diffusion model} generates the target latent conditioned on the observed latent (details in Sec.~\ref{sec:latent-diffusion}). The forward corruption uses the observed representation as its mean-reversion reference, and iterative latent recovery produces a continuous estimate for triple decoding.

\subsection{Relation-aware Residual-Quantized KG AutoEncoder}\label{sec:KG-autoencoder}
The goal of \modelname's first stage (Stage I) is to learn a compact and reconstructive latent space for KGs. To better preserve relation-specific adjacency patterns, we design a relation-aware residual-quantized KG autoencoder.

\subsubsection{Variable-Density KG Construction}
Stage I is trained as an autoencoder, where the input graph itself serves as the reconstruction target. To expose the autoencoder to KGs with different sparsity levels, we randomly sparsify the training graph at each iteration. For each training iteration, we sample a ratio $\rho \sim \mathcal{U}(0, 1),$ and construct a masked KG $\mathcal{G}_{\rho}=(\mathcal{E},\mathcal{R},\mathcal{T}_{\rho})$ by independently keeping each positive triple in the training graph with probability $\rho$. The autoencoder takes $\mathcal{G}_{\rho}$ as input and is trained to reconstruct the input.  

\subsubsection{Relation-aware Graph Encoder}
For each head entity $h$ and relation $r$, we define a tail indicator vector $ \mathbf{a}_{h,r}\in\{0,1\}^{|\mathcal{E}|},$ where the $t$-th entry is $1$ if $(h,r,t)$ is in the KG and $0$ otherwise. 
The relation-aware graph encoder maps each sparse tail indicator vector into a dense latent embedding: $ \mathbf{x}_{h,r} = W_a \mathbf{a}_{h,r},$ where $W_a$ is a projection matrix. 
For each head entity $h$, we collect its relation-specific latent as $\mathbf{X}_{h}$:
{\small 
\begin{equation}
    \mathbf{X}_{h} =[\mathbf{x}_{h,1},\mathbf{x}_{h,2},\ldots,\mathbf{x}_{h,|\mathcal{R}|}].
\end{equation}}
\noindent To model dependencies among different relational views, we propose relational view attention:
{\small 
\begin{equation}
    \widetilde{\mathbf{X}}_{h} = \mathrm{Attn}_{R}(\mathbf{X}_{h}),
\end{equation}}
\noindent where attention is performed along the relation dimension. 
The output is combined with the original projected representation through a residual connection:
$  \mathbf{X}_{h}  =  \mathbf{X}_{h}  +  \widetilde{\mathbf{X}}_{h}.$
Stacking all head entities gives the encoder output
$ \mathbf{X} = {E_{\phi}}(\mathcal{G}_{\rho}) 
\in \mathbb{R}^{|\mathcal{E}|\times|\mathcal{R}|\times d},$ where $E_{\phi}$ is the encoder.

\subsubsection{Chunked Residual Vector Quantization}

The continuous encoder output is further discretized by a chunked residual vector quantizer.  Specifically, we decompose each latent embedding into $K$ chunks:
$
    \mathbf{x}_{h,r}
    =
    [
    \mathbf{x}_{h,r}^{(1)},
    \mathbf{x}_{h,r}^{(2)},
    \ldots,
    \mathbf{x}_{h,r}^{(K)}
    ].
$
For each chunk, we apply residual vector quantization with $L_c$ codebooks. 
Let $\mathbf{r}^{(k,1)}=\mathbf{x}_{h,r}^{(k)}$ be the initial residual of the $k$-th chunk. 
At the $\ell$-th residual quantization level, the nearest codeword is selected as
{\small
 \begin{equation}
    c^{(k,\ell)}
    =
    \arg\min_j
    \left\|
    \mathbf{r}^{(k,\ell)}
    -
    \mathbf{e}_{j}^{(k,\ell)}
    \right\|_2^2,
\end{equation} }
\noindent where $\mathbf{e}_{j}^{(k,\ell)}$ is the $j$-th codeword in the $\ell$-th codebook of the $k$-th chunk. 
The residual is then updated by
{\small
\begin{equation}
    \mathbf{r}^{(k,\ell+1)}
    =
    \mathbf{r}^{(k,\ell)}
    -
    \mathbf{e}_{c^{(k,\ell)}}^{(k,\ell)}.
\end{equation}}
The quantized representation of the $k$-th chunk is
\begin{small}
\begin{equation}
    \mathbf{q}_{h,r}^{(k)}
    =
    \sum_{\ell=1}^{L_c}
    \mathbf{e}_{c^{(k,\ell)}}^{(k,\ell)}.
\end{equation}    
\end{small}

\noindent Finally, the quantized latent embedding is obtained by concatenating all quantized chunks:
$
    \mathbf{z}_{\mathrm{q},h,r}
    =
    [
    \mathbf{q}_{h,r}^{(1)},
    \mathbf{q}_{h,r}^{(2)},
    \ldots,
    \mathbf{q}_{h,r}^{(K)}
    ].
$ We adopt the straight-through estimator for the non-differentiable nearest-neighbor assignment:
\begin{equation}
    \mathbf{z}_{h,r}
    =
    \mathbf{x}_{h,r}
    +
    \mathrm{sg}\!\left(
    \mathbf{z}_{\mathrm{q},h,r}-\mathbf{x}_{h,r}
    \right),
\end{equation}
\noindent which equals $\mathbf{z}_{\mathrm{q},h,r}$ in the forward pass while propagating gradients to $\mathbf{x}_{h,r}$ in the backward pass, $\mathbf{z}_{h,r}$ is used as the decoder input and $\rm sg(\cdot)$ is the stop-gradient operator.
The chunk-level VQ training objective is
\begin{small}
\begin{equation}
\begin{aligned}
    \mathcal{L}_{\mathrm{VQ}} = & \frac{1}{|\Omega|K} \sum_{(h,r)\in\Omega}\sum_{k=1}^{K} \biggl[  \left\| \mathrm{sg}(\mathbf{x}_{h,r}^{(k)}) - \mathbf{q}_{h,r}^{(k)} \right\|_2^2 \\
    & + \lambda_m \left\| \mathbf{x}_{h,r}^{(k)} - \mathrm{sg}(\mathbf{q}_{h,r}^{(k)}) \right\|_2^2 \biggr],
\end{aligned}
\end{equation}
\end{small}

\noindent where $\Omega$ denotes the set of head-relation pairs, $K$ is the number of chunks, and $\lambda_m$ is the commitment weight.

\subsubsection{Factorized Triple Decoder}
The decoder reconstructs triples from the quantized latent embeddings.  For a triple $(h,r,t)$, the factorized scoring function is defined as
\begin{equation}
    s(h,r,t)=
    \left\langle
    U(\mathbf{z}_{h,r}),
    \mathbf{e}_t
    \right\rangle
    +
    b_r,
\end{equation}
where $\mathbf{z}_{h,r}$ is the quantized latent representation of the head-relation pair $(h,r)$, $\mathbf{e}_t$ is the tail entity embedding, $U(\cdot)$ is the projection function, and $b_r$ is relation bias. 
The probability of a triple being positive is then $p(h,r,t)=  \sigma(s(h,r,t)).$ 

\subsubsection{Training Objective}

We train the relation-aware residual-quantized KG autoencoder with negative sampling.
For each positive triple $(h,r,t)\in\mathcal{T}_{{\rho}}$, we sample $K_{\mathrm{neg}}$ negative triples by corrupting the tail entity:
\begin{small}
\begin{equation}
    \mathcal{N}(h,r)
    =
    \{(h,r,t_i^-)\}_{i=1}^{K_{\mathrm{neg}}},
    \quad
    (h,r,t_i^-)\notin \mathcal{T}_{{\rho}}.
\end{equation}    
\end{small}

\noindent The reconstruction loss is
\begin{small}
    \begin{equation}
\begin{aligned}
    \mathcal{L}_{\mathrm{rec}} = 
    &- \sum_{(h,r,t)\in\mathcal{T}_{{\rho}}} \log \sigma(s(h,r,t)) \\
    &- \lambda_{\mathrm{neg}} \sum_{(h,r,t^-)\in\mathcal{N}} \log \sigma(-s(h,r,t^-)).
\end{aligned}
\end{equation}
\end{small}

\noindent The overall Stage-I objective is
\begin{equation}
    \mathcal{L}_{\mathrm{AE}}
    =
    \mathcal{L}_{\mathrm{rec}}
    +
    \lambda_{\mathrm{VQ}}
    \mathcal{L}_{\mathrm{VQ}}.
\end{equation}
Here, $\mathcal{L}_{\mathrm{rec}}$ encourages the quantized latent to preserve information needed for triple reconstruction, while $\mathcal{L}_{\mathrm{VQ}}$ aligns the continuous encoder outputs with the discrete residual codebook space. 

\paragraph{Reconstruction Fidelity Metrics}
We report reconstruction recall ($RecRecall$), precision ($RecPrecision$), and their harmonic mean ($RF_1$) as diagnostics of how faithfully the learned latent space reconstructs input triples. Detailed definitions are provided in the Appendix.

\subsection{Observation-Anchored Latent Graph Diffusion}\label{sec:latent-diffusion}
After the knowledge graph autoencoder is trained, we use it to map both observed and target KGs into the latent space. 
The second stage (Stage II) of \modelname ~is to learn a conditional diffusion model $p_{\theta}(\mathbf{Z}_{0}\mid \mathbf{Z}_{\mathrm{obs}}),$ where $\mathbf{Z}_{\mathrm{obs}}$ is the latent representation of the observed KG and $\mathbf{Z}_{0}$ is the latent representation of the target KG. 

\subsubsection{Latent Priors Construction}
During Stage-II training, we construct paired observed-target KGs from the training graph. 
Specifically, we sample $\rho_{\mathrm{tar}}\sim\mathcal{U}(\rho_{\rm min},1.0)$ and retain each training triple
with probability $\rho_{\mathrm{tar}}$ to construct $\mathcal{G}_{\mathrm{tar}}$. Then, we obtain $\mathcal{G}_{\mathrm{obs}}$ by retaining each triple in $\mathcal{G}_{\mathrm{tar}}$ with probability $\rho_{\mathrm{obs}}$, ensuring $\mathcal{T}_{\mathrm{obs}} \subseteq\mathcal{T}_{\mathrm{tar}} \subseteq\mathcal{T}_{\mathrm{train}}$.
With the trained encoder and quantizer in Stage I, we have

{\small
\begin{equation}
    \mathbf{Z}_{\mathrm{obs}}
    =
    Q_{\phi}(E_{\phi}(\mathcal{G}_{obs})),
    \quad
    \mathbf{Z}_{0}
    =
    Q_{\phi}(E_{\phi}(\mathcal{G}_{tar})),
\end{equation}}

\noindent where $E_{\phi}$ is the relation-aware graph encoder and $Q_{\phi}$ is the chunked residual quantizer. 
The autoencoder parameters are frozen during diffusion training.
% , so the second stage learns generation within a fixed latent space
\subsubsection{Mean-Reverting Forward Process}
Standard diffusion models perturb data toward an unconditional Gaussian prior. 
However, for TSP, the observed graph already provides a strong structural anchor. 
Therefore, we adopt a mean-reverting forward process following prior restoration diffusion models~\cite{luo2023imageirsde} and specialize its endpoint and conditioning structure for TSP.
Let $\mathbf{Z}_{0}$ denote the clean latent representation of the target KG and $\mathbf{Z}_{\mathrm{obs}}$ denote the latent representation of the observed KG. 
We define a linear variance schedule $\{\beta_s\}_{s=1}^{S}$, where $S$ is the total number of diffusion steps. 
Following DDPM, we set
\begin{small}
\begin{equation}
    \alpha_s = 1-\beta_s,
    \qquad
    \bar{\alpha}_s = \prod_{i=1}^{s}\alpha_i .
\end{equation}
\end{small}

\noindent Here, $\bar{\alpha}_s$ controls the signal-to-noise ratio at diffusion step $s$. 
The mean-reverting forward process is defined as
\begin{small}
 \begin{equation}
\label{eq:deltadiff}
    q(\mathbf{Z}_{s}\mid \mathbf{Z}_{0},\mathbf{Z}_{\mathrm{obs}})
    =
    \mathcal{N}
    \left(
    \mathbf{Z}_{\mathrm{obs}}
    +
    \sqrt{\bar{\alpha}_{s}}
    (\mathbf{Z}_{0}-\mathbf{Z}_{\mathrm{obs}}),
    (1-\bar{\alpha}_{s})\mathbf{I}
    \right).
\end{equation}   
\end{small}

\noindent When $s$ is small, $\bar{\alpha}_s$ is close to $1$, and the noisy latent remains close to the target latent. When $s$ approaches $S$, $\bar{\alpha}_s$ becomes close to $0$, and the noisy latent is mainly centered around the observed-graph latent with Gaussian noise.
Equivalently, by defining the missing-information residual \(\Delta_0 = Z_0 - Z_{\text{obs}}\) and \(\Delta_s = Z_s - Z_{\text{obs}}\), Eq.~\ref{eq:deltadiff} reduces to the standard DDPM marginal \(q(\Delta_s|\Delta_0) = \mathcal{N}(\sqrt{\bar{\alpha}_s}\Delta_0, (1 - \bar{\alpha}_s)I)\). Therefore, observation-anchored diffusion performs residual denoising in the latent representation. We instantiate mean-reverting diffusion on the proposed relation-aware latent space and couple it with observation conditioning and dual-view denoising for TSP.

\subsubsection{Dual-view Latent Graph Denoiser}
The denoising network predicts the clean target latent from the noisy latent $\mathbf{Z}_{s}$, the observed-graph latent $\mathbf{Z}_{\mathrm{obs}}$, and the diffusion step $s$:
\begin{equation}
{\small
\widehat{\mathbf{Z}}_{0}
=
W_{\mathrm{out}}\,
\mathrm{DLGD}_{L}
\!\left(
\mathrm{FiLM}_{s}
\!\left(
W_{\mathrm{in}}
[\mathbf{Z}_{s};\mathbf{Z}_{\mathrm{obs}}]
\right)
\right).
}
\end{equation}

\noindent Here, $[\cdot\,;\cdot]$ denotes feature-wise concatenation, $W_{\mathrm{in}}$ and $W_{\mathrm{out}}$ are learnable input and output projections, respectively, and $\mathrm{DLGD}_{L}$ denotes a stack of $L$ Dual-view Latent Graph Denoiser (DLGD) blocks. The diffusion step is incorporated through FiLM conditioning, i.e., $\mathrm{FiLM}_{s}(\mathbf{\cdot})$. To efficiently capture dependencies along both the relation and entity axes, each block performs a dual-view linear attention:
\begin{equation}
{\small
\begin{aligned}
    \mathbf{H}_{R}
    &=
    \mathbf{H}
    +
    \mathrm{LinAttn}_{R}
    \bigl(\mathrm{LN}(\mathbf{H})\bigr),\\
    \mathbf{H}_{N}
    &=
    \mathbf{H}_{R}
    +
    \mathrm{LinAttn}_{N}
    \bigl(\mathrm{LN}(\mathbf{H}_{R})\bigr),\\
    \mathbf{H}^{\mathrm{out}}
    &=
    \mathbf{H}_{N}
    +
    \mathrm{FFN}
    \bigl(\mathrm{LN}(\mathbf{H}_{N})\bigr).
\end{aligned}
}
\end{equation}

\noindent Here, $\mathbf{H}$ and $\mathbf{H}^{\mathrm{out}}$ are the block input and output, while $\mathbf{H}_{R}$ and $\mathbf{H}_{N}$ are intermediate representations after relation-wise and node-wise linear attention, respectively. $\mathrm{LN}$ denotes layer normalization and $\mathrm{FFN}$ denotes a feed-forward network. Specifically, $\mathrm{LinAttn}_{R}$ exchanges information across relations associated with the same head entity, whereas $\mathrm{LinAttn}_{N}$ aggregates information across head entities under the same relation. Their combination enables efficient dual-view dependency modeling over the unified latent representation.

\subsubsection{Diffusion Training Objective}

During training, we randomly sample a diffusion step 
$s\sim\mathrm{Uniform}(\{1,\ldots,S\})$ 
and Gaussian noise 
$\boldsymbol{\epsilon}\sim\mathcal{N}(\mathbf{0},\mathbf{I})$. 
The noisy latent $\mathbf{Z}_{s}$ is constructed by the mean-reverting forward process. 
The denoiser is trained to predict the clean target latent $\mathbf{Z}_{0}$:
\begin{small}
\begin{equation}
    \mathcal{L}_{\mathrm{diff}}
    =
    \mathbb{E}_{s,\boldsymbol{\epsilon}}
    \left[
    \left\|
    f_{\theta}(\mathbf{Z}_{s},\mathbf{Z}_{\mathrm{obs}},s)
    -
    \mathbf{Z}_{0}
    \right\|_2^2
    \right].
\end{equation}    
\end{small}

\subsubsection{Inference and Triple Set Decoding}

At inference, the observed KG $\mathcal{G}_{\mathrm{obs}}$ is encoded into
$\mathbf{Z}_{\mathrm{obs}}$. Since $\bar{\alpha}_{S}\approx 0$, we initialize
$\mathbf{Z}_{S}=\mathbf{Z}_{\mathrm{obs}}+\boldsymbol{\epsilon}$, where
$\boldsymbol{\epsilon}\sim\mathcal{N}(\mathbf{0},\mathbf{I})$. At each reverse
step $s=S,\ldots,1$, the denoiser predicts
$\widehat{\mathbf{Z}}_{0}
=f_{\theta}(\mathbf{Z}_{s},\mathbf{Z}_{\mathrm{obs}},s)$, followed by
DDPM sampling:
\begin{equation}
{\small
\begin{aligned}
\boldsymbol{\mu}_{s}
={}&
\mathbf{Z}_{\mathrm{obs}}
+
\frac{\sqrt{\bar{\alpha}_{s-1}}\beta_s}
     {1-\bar{\alpha}_{s}}
\bigl(\widehat{\mathbf{Z}}_{0}-\mathbf{Z}_{\mathrm{obs}}\bigr)\\
&+
\frac{\sqrt{\alpha_s}(1-\bar{\alpha}_{s-1})}
     {1-\bar{\alpha}_{s}}
\bigl(\mathbf{Z}_{s}-\mathbf{Z}_{\mathrm{obs}}\bigr),\\
\mathbf{Z}_{s-1}
={}&
\boldsymbol{\mu}_{s}
+
\mathbb{I}[s>1]
\sqrt{\tilde{\beta}_{s}}\boldsymbol{\xi},
\qquad
\boldsymbol{\xi}\sim\mathcal{N}(\mathbf{0},\mathbf{I}),
\end{aligned}
}
\end{equation}
where $\tilde{\beta}_{s} =\beta_s(1-\bar{\alpha}_{s-1})/(1-\bar{\alpha}_{s})$ is the posterior variance, and $\mathbb{I}[s>1]$ removes the stochastic noise at the final reverse step.
After the final reverse step, we obtain the generated latent $\widehat{\mathbf{Z}}_0$. We decode this representation without an additional quantization step. Since the factorized decoder operates on latent vectors instead of discrete code indices, it maps $\widehat{\mathbf{Z}}_0$ to candidate triple scores as:
\begin{equation}
s(h,r,t)=\left\langle U(\hat{\mathbf{z}}_{0,h,r}),\mathbf{e}_t\right\rangle+b_r.
\end{equation}
Triples already contained in $\mathcal{T}_{\mathrm{obs}}$ are filtered out.
Following the standard TSP evaluation protocol, a single validation-selected score threshold is applied to the remaining candidates to obtain the final predicted missing triple set $\hat{\mathcal{T}}_{\mathrm{miss}}$. The same threshold-selection procedure is applied to all compared methods to ensure a consistent evaluation.

% This output rule imposes no additional logical constraints on the selected triples.

\paragraph{Method Details.} Detailed dual-view linear attention and complexity analysis are provided in the Appendix.

\section{Experiments}
\subsection{Experimental Setup}
\subsubsection{Datasets}
We evaluate our method on three common benchmark datasets for TSP: Wiki79k, Wiki143k, and CFamily. 
Table~\ref{tab:datasets} lists the entity, relation, and triple counts for each dataset.
We strictly use the original data partitions and definitions provided by GPHT~\cite{zhang2024gpht}, which guarantees that our results are comparable with existing methods.

\begin{table}[ht]
    \centering
    % \vspace{-5pt}
      \caption{Statistics of datasets in experiments.}
      
    \label{tab:datasets}
     \scalebox{0.8}{%
    \begin{tabular}{lccccc}
        \toprule
        Datasets & Entity & Relation & Train & Valid & Test \\
        \midrule
        Wiki79k & 7,983 & 85 & 57,033 & 6,337 & 15,843 \\
        Wiki143k & 13,928 & 109  & 103,415 & 11,490 & 28,727  \\
        CFamily & 2,378 & 12  & 16,549 & 1,839 & 4,598  \\
        \bottomrule
    \end{tabular}
    }   
\end{table}

% \subsubsection{Baselines}
% We select both link prediction methods and TSP methods as baselines. The representative link prediction methods, including TensorLog, HAKE, PairRE, RGCN, CompGCN, AstarNet and ULTRA. The latest TSP methods, such as GPHT and LLMTSP, are also included.

\subsubsection{Evaluation Metrics}
To ensure a fair and comprehensive comparison, we adopt the evaluation metrics proposed for the TSP task in GPHT~\cite{zhang2024gpht}, including Joint Precision 
($\textit{JPrecision}$), Square-root Test Recall ($\textit{STRecall}$), and TSP Score ($F_{\rm TSP}$). Because $STRecall$ is the square root of conventional test recall, $F_{\rm TSP}$ differs from conventional $F_1$.  More details are in the Appendix.

\subsection{Main Results}
Table~\ref{tab:owa} and Table~\ref{tab:cwa} report the main results on all datasets. For \modelname{}, we report the mean and standard deviation over three independent runs.  We take $F_{\rm{TSP}}$ as the primary comparison metric because it balances the precision of the predicted triple set and the recovery of ground-truth triples. 
Therefore, we rank different methods mainly according to $F_{{\rm TSP}}$, while using ${JPrecision}$ and ${STRecall}$ to interpret the precision-recall behavior of each method. We also report the standard $F_1$, reconstructed from the  reported mean triple counts.
Overall, \modelname~ achieves the best $F_{\rm TSP}$ on all three datasets, outperforming the strongest baseline by absolute 5.4 percentage points on average.

\begin{table*}[ht]
    \centering
    \caption{The results on Wiki79k and Wiki143k, where
    $\mathcal T^{\rm POWA+}_{\rm pred}$ is the number of correctly predicted
    test triples, and $\mathcal T^{\rm POWA}_{\rm pred}$ is the triple set
    that can be judged under the RS-POWA assumption.
    The best results are in bold, and the second-best results are underlined.
    $F_{\rm TSP}$ is the primary comparison metric.
    $\dagger$: Standard $F_1$ under RS-POWA. }
    \label{tab:owa}
    \resizebox{\textwidth}{!}{%
    \begin{tabular}{l|l|ccc|cc>{\columncolor{gray!10}}cc|c>{\columncolor{gray!10}}c}
        \toprule
        Datasets & Models
        & \multicolumn{3}{c|}{Number of Triples in}
        & \multicolumn{4}{c|}{RS-POWA Metrics}
        & \multicolumn{2}{c}{CWA Metrics} \\
        \midrule
        & & ${\mathcal T}_{\mathrm{pred}}$
        & ${\mathcal T}^{\mathrm{POWA}}_{\mathrm{pred}}$
        & ${\mathcal T}^{\mathrm{POWA}+}_{\mathrm{pred}}$
        & $JPrecision$ & $STRecall$ & $F_{\rm TSP}$ & $F_1^\dagger$
        & $JPrecision$ & $F_{\rm TSP}$ \\
        \midrule

        \multirow{8}{*}{Wiki79k}
        & HAKE~\cite{zhang2020hake}
        & 125$\pm$87 & 119$\pm$83 & 30$\pm$7
        & 0.246$\pm$14.8\% & 0.044$\pm$0.4\% & 0.075$\pm$0.3\%
        & 0.004
        & 0.240 & 0.074 \\

        & PairRE~\cite{chao2020pairre}
        & 29742$\pm$105 & 10217$\pm$51 & 2901$\pm$7
        & 0.191$\pm$0.1\% & 0.428$\pm$0.0\% & 0.264$\pm$0.2\%
        & 0.187
        & 0.097 & 0.158 \\

        & AstarNet~\cite{zhu2023astarnet}
        & 5689$\pm$287 & 1750$\pm$116 & 1328$\pm$52
        & 0.496$\pm$0.8\% & 0.289$\pm$0.5\% & 0.365$\pm$0.2\%
        & 0.143
        & 0.233 & 0.258 \\

        & ULTRA~\cite{2024ultra}
        & 9743$\pm$0 & 2945$\pm$0 & 1062$\pm$0
        & 0.235$\pm$0.0\% & 0.258$\pm$0.0\% & 0.246$\pm$0.0\%
        & 0.104
        & 0.109 & 0.153 \\

        & GPHT(HAKE)~\cite{zhang2024gpht}
        & 209$\pm$47 & 191$\pm$47 & 44$\pm$8
        & 0.220$\pm$8.0\% & 0.053$\pm$0.4\% & 0.085$\pm$1.1\%
        & 0.006
        & 0.210 & 0.084 \\

        & GPHT(PairRE)~\cite{zhang2024gpht}
        & 12392$\pm$3813 & 5866$\pm$1262 & 2018$\pm$332
        & 0.253$\pm$1.6\% & 0.357$\pm$2.8\% & 0.296$\pm$0.4\%
        & 0.170
        & 0.162 & 0.224 \\

        & SMuRT~\cite{yuan2026smurttsp} & 13450$\pm$0 & 3197$\pm$0 & 1245$\pm$0 & \text{0.241$\pm$0.0\%} & 0.280$\pm$0.0\% & \text{0.259$\pm$0.0\%} & 0.118 
        & 0.093 & 0.139  \\
        
        & DiffTSP~\cite{guan2026difftsp}
        & 7472$\pm$103 & 4355$\pm$51 & 3472$\pm$87
        & \underline{0.630$\pm$1.2\%}
        & \underline{0.468$\pm$0.5\%}
        & \underline{0.537$\pm$0.5\%}
        & \underline{0.325}
        & \underline{0.464} & \underline{0.466} \\

        \cmidrule{2-11}
        & \modelname
        & 7384$\pm$52 & 4871$\pm$34 & 4060$\pm$23
        & \textbf{0.692$\pm$0.5\%}
        & \textbf{0.506$\pm$0.1\%}
        & \textbf{0.585$\pm$0.3\%}
        & \textbf{0.374}
        & \textbf{0.550} & \textbf{0.527} \\

        \midrule

        \multirow{8}{*}{Wiki143k}
        & HAKE~\cite{zhang2020hake}
        & 22215$\pm$1283 & 6182$\pm$43 & 3044$\pm$72
        & 0.315$\pm$0.2\% & 0.326$\pm$0.3\% & 0.32$\pm$0.3\%
        & 0.159
        & 0.137 & 0.193 \\

        & PairRE~\cite{chao2020pairre}
        & 19228$\pm$2075 & 3313$\pm$722 & 1191$\pm$77
        & 0.211$\pm$3.0\% & 0.204$\pm$0.7\% & 0.207$\pm$1.0\%
        & 0.070
        & 0.061 & 0.095 \\

        & AstarNet~\cite{zhu2023astarnet}
        & 6001$\pm$231 & 2589$\pm$129 & 2275$\pm$31
        & \underline{0.629$\pm$0.8\%}
        & 0.281$\pm$0.2\% & 0.389$\pm$0.2\%
        & 0.140
        & 0.379 & 0.323 \\

        & ULTRA~\cite{2024ultra}
        & 3844$\pm$0 & 1349$\pm$0 & 692$\pm$0
        & 0.346$\pm$0.0\% & 0.155$\pm$0.0\% & 0.214$\pm$0.0\%
        & 0.045
        & 0.180 & 0.166 \\

        & GPHT(HAKE)~\cite{zhang2024gpht}
        & 17702$\pm$7935 & 4709$\pm$60 & 2700$\pm$681
        & 0.363$\pm$4.8\% & 0.307$\pm$4.2\% & 0.333$\pm$4.6\%
        & 0.150
        & 0.152 & 0.204 \\

        & GPHT(PairRE)~\cite{zhang2024gpht}
        & 3011$\pm$233 & 1954$\pm$191 & 909$\pm$34
        & 0.384$\pm$1.9\% & 0.178$\pm$0.3\% & 0.243$\pm$0.1\%
        & 0.059
        & 0.302 & 0.224 \\

        & DiffTSP~\cite{guan2026difftsp}
        & 10162$\pm$530 & 6804$\pm$275 & 4543$\pm$105
        & 0.557$\pm$0.3\%
        & \underline{0.397$\pm$0.4\%}
        & \underline{0.464$\pm$0.6\%}
        & \underline{0.246}
        & \underline{0.447} & \underline{0.420} \\

        \cmidrule{2-11}
        & \modelname
        & 11211$\pm$57 & 5748$\pm$20 & 5149$\pm$42
        & \textbf{0.677$\pm$0.5\%}
        & \textbf{0.423$\pm$0.2\%}
        & \textbf{0.521$\pm$0.1\%}
        & \textbf{0.283}
        & \textbf{0.459} & \textbf{0.440} \\

        \bottomrule
    \end{tabular}%
    }
\end{table*}
\begin{table*}[ht]
    \centering
    \caption{The results on the CFamily dataset, where
    ${\mathcal T}^{\mathrm{CWA}+}_{\mathrm{pred}}$ is the set of correctly
    predicted test triples.
    The best results are in bold, and the second-best results are underlined.
    $F_{\rm TSP}$ is the primary comparison metric.
    $\dagger$: Standard $F_1$ under CWA.}
    \label{tab:cwa}
    \scalebox{0.9}{%
    \begin{tabular}{l|ccc|cc>{\columncolor{gray!10}}cc}
        \toprule
        \multicolumn{1}{c|}{Models}
        & \multicolumn{3}{c|}{Number of Triples in}
        & \multicolumn{4}{c}{CWA Metrics} \\
        \midrule
        & ${\mathcal T}_{\mathrm{pred}}$
        & ${\mathcal T}^{\mathrm{CWA}}_{\mathrm{pred}}$
        & ${\mathcal T}^{\mathrm{CWA}+}_{\mathrm{pred}}$
        & $JPrecision$ & $STRecall$ & $F_{\rm TSP}$ & $F_1^\dagger$ \\
        \midrule

        HAKE~\cite{zhang2020hake}
        & 3186$\pm$543 & 3186$\pm$543 & 1788$\pm$321
        & 0.561$\pm$1.8\% & 0.624$\pm$5.9\% & 0.591$\pm$3.1\%
        & 0.459 \\

        PairRE~\cite{chao2020pairre}
        & 5732$\pm$2062 & 5732$\pm$2062 & 1747$\pm$694
        & 0.305$\pm$1.8\% & 0.616$\pm$13.7\% & 0.408$\pm$4.9\%
        & 0.338 \\

        RGCN~\cite{2018rgcn}
        & 30608$\pm$3443 & 30608$\pm$3443 & 3026$\pm$137
        & 0.099$\pm$6.0\% & \textbf{0.811$\pm$0.5\%} & 0.176$\pm$5.2\%
        & 0.172 \\

        CompGCN~\cite{CompGCN}
        & 38931$\pm$3745 & 38931$\pm$3745 & 2599$\pm$47
        & 0.067$\pm$4.8\% & \underline{0.752$\pm$0.3\%} & 0.123$\pm$1.7\%
        & 0.119 \\

        AstarNet~\cite{zhu2023astarnet}
        & 2075$\pm$360 & 2075$\pm$360 & 911$\pm$187
        & 0.439$\pm$1.7\% & 0.445$\pm$4.6\% & 0.442$\pm$3.3\%
        & 0.273 \\

        ULTRA~\cite{2024ultra}
        & 32761$\pm$0 & 32761$\pm$0 & 763$\pm$0
        & 0.023$\pm$0.0\% & 0.407$\pm$0.0\% & 0.044$\pm$0.0\%
        & 0.041 \\

        GPHT(HAKE)~\cite{zhang2024gpht}
        & 1896$\pm$149 & 1896$\pm$149 & 1222$\pm$58
        & 0.645$\pm$2.6\% & 0.516$\pm$1.2\% & 0.573$\pm$0.4\%
        & 0.376 \\

        GPHT(PairRE)~\cite{zhang2024gpht}
        & 3739$\pm$593 & 3739$\pm$593 & 1471$\pm$187
        & 0.393$\pm$1.3\% & 0.566$\pm$3.4\% & 0.464$\pm$0.7\%
        & 0.353 \\

        LLMTSP(GPT-3.5)~\cite{yuan2024llmtsp}
        & 3403$\pm$158 & 3403$\pm$158 & 96$\pm$12
        & 0.028$\pm$0.3\% & 0.145$\pm$0.8\% & 0.047$\pm$0.4\%
        & 0.024 \\

        LLMTSP(GPT-4o)~\cite{yuan2024llmtsp}
        & 1276$\pm$147 & 1276$\pm$147 & 179$\pm$17
        & 0.140$\pm$0.5\% & 0.197$\pm$0.9\% & 0.164$\pm$0.3\%
        & 0.061 \\

        SMuRT~\cite{yuan2026smurttsp} & 2358$\pm$0 & 2358$\pm$0 &657$\pm$0 & 0.279$\pm$0.0\% & 0.378$\pm$0.0\% & 0.321$\pm$0.0\% & 0.189\\
        
        DiffTSP~\cite{guan2026difftsp}
        & 2453$\pm$35 & 2453$\pm$35 & 1657$\pm$18
        & \underline{0.675$\pm$0.2\%}
        & 0.600$\pm$0.3\%
        & \underline{0.635$\pm$0.2\%}
        & \underline{0.470} \\

        \midrule
        \modelname
        & 3299$\pm$60 & 3299$\pm$60 & 2252$\pm$39
        & \textbf{0.683$\pm$0.5\%}
        & \text{0.700$\pm$0.5\%}
        & \textbf{0.691$\pm$0.2\%}
        & \textbf{0.570} \\

        \bottomrule
    \end{tabular}%
    }
\end{table*}

\subsection{Ablation Study}
Table~\ref{tab:ablation} reports the ablation results on three datasets using $F_{\mathrm{TSP}}$ as the primary metric. The full \modelname{} achieves the best performance on all datasets, demonstrating the effectiveness of the overall design.
\textit{\modelname{-Subgraph}} retains the same modules as \modelname{}, but partitions the KG into local subgraphs like GPHT.  The performance drop provides a controlled comparison that supports the benefit of operating in a unified graph latent space. 
For Stage I, \textit{w/o CRVQ} replaces CRVQ with a codebook parameter matched VQ-VAE while keeping the encoder, decoder, latent dimensionality, training protocol, and total number of learnable codebook parameters comparable. Its substantial degradation is therefore not attributable to a smaller parameter budget. VQ-VAE exhibits poor Stage-I reconstruction fidelity and provides an inadequate latent space for subsequent diffusion. Detailed results are provided in the Appendix. 
\textit{w/o Quantization} removes the quantization module and directly uses the continuous encoder output as the latent representation. The degradation of \textit{w/o Quantization} suggests that unconstrained continuous latents are less suitable for diffusion-based TSP, supporting the role of quantization in constructing a structured latent space. Removing the relational-view attention (\textit{w/o Relational view Attention}) or replacing the factorized decoder (\textit{w/o Factorized Decoder}) with MLPs also leads to clear performance drops. 
\textit{w/o Stage II} directly decodes the observed latent. Its poor performance shows that our proposed Stage-II module provides the conditional latent transformation required for TSP. 

For Stage II, standard Gaussian diffusion decreases the average score to 0.542, confirming the importance of the mean-reverting diffusion process that anchors generation on the observed KG. \textit{w/o Iterative Diffusion} replaces diffusion with a one-step predictor while retaining the same denoiser backbone, the resulting performance drop demonstrates the benefit of progressive latent diffusion refinement. 
We further compare \modelname{} with a deterministic iterative refinement baseline using the same denoiser backbone and refinement steps. Unlike diffusion, this baseline learns to progressively update the latent representation without a noise-based forward process. Its lower performance suggests that the benefit of \modelname{} extends beyond repeated latent refinement.
In addition, removing either relation-wise (\textit{w/o Relation-wise Linear Attention}) or node-wise linear attention (\textit{w/o Node-wise Linear Attention}) consistently hurts performance, suggesting that both relation-level and node-level dependencies contribute to effective latent generation.

\paragraph{Construction of \modelname{}-Subgraph.} We use exactly the graph partitioning procedure and partition hyperparameters of GPHT for each dataset. The same \modelname{} parameters are shared across all subgraphs. Predictions produced by multiple overlapping subgraphs are merged by union, while triples whose endpoints never co-occur in a subgraph cannot be generated. The prediction threshold is independently selected on the validation set for this variant. All hyperparameters are re-tuned using the same search space and validation budget as \modelname{}.
%mean-re改为Observation-Anchored Residual Diff

\begin{table}[t]
\centering
\caption{Ablation study of \modelname{} on three datasets. We report $F_{\mathrm{TSP}}$ for TSP evaluation. }
% Full results with $\textit{JPrecision}$ and $\textit{STRecall}$ are provided in the Appendix.}
\label{tab:ablation}
\resizebox{1\linewidth}{!}{
\begin{tabular}{lcccc}
\toprule
Variant 
& Wiki79k 
& Wiki143k 
& CFamily 
& Avg. \\
\midrule
~\modelname 
& \textbf{0.585} & \textbf{0.521} & \textbf{0.691} & \textbf{0.599} \\
~\modelname{-Subgraph} 
& 0.539 & 0.472 & 0.653 & 0.555 \\
\midrule
w/o Relational view Attention
& 0.576 & 0.495 & 0.674 & 0.582 \\
w/o CRVQ (VQ-VAE)
& 0.030 & 0.075 & 0.455 & 0.187 \\
w/o Quantization (Vanilla AE)
& 0.498 & 0.414 & 0.332 & 0.415 \\
w/o Factorized Decoder 
& 0.557 & 0.481 & 0.659 & 0.566 \\
\midrule
w/o Stage II (Direct Decoding)
& 0.277 & 0.221 & 0.351 & 0.283 \\
w/o Iterative Diffusion
& 0.458 & 0.442 & 0.518 & 0.473 \\
Deterministic Iterative Refinement
& 0.501 & 0.471 & 0.615 & 0.529 \\
Standard Gaussian Diffusion 
& 0.513 & 0.476 & 0.637 & 0.542 \\
w/o Relation-wise Linear Attention 
& 0.538 & 0.493 & 0.554 & 0.528 \\
w/o Node-wise Linear Attention 
& 0.546 & 0.508 & 0.602 & 0.552 \\
\bottomrule
\end{tabular}
}
\end{table}

\begin{figure}
    \centering
    \includegraphics[width=0.9\linewidth]{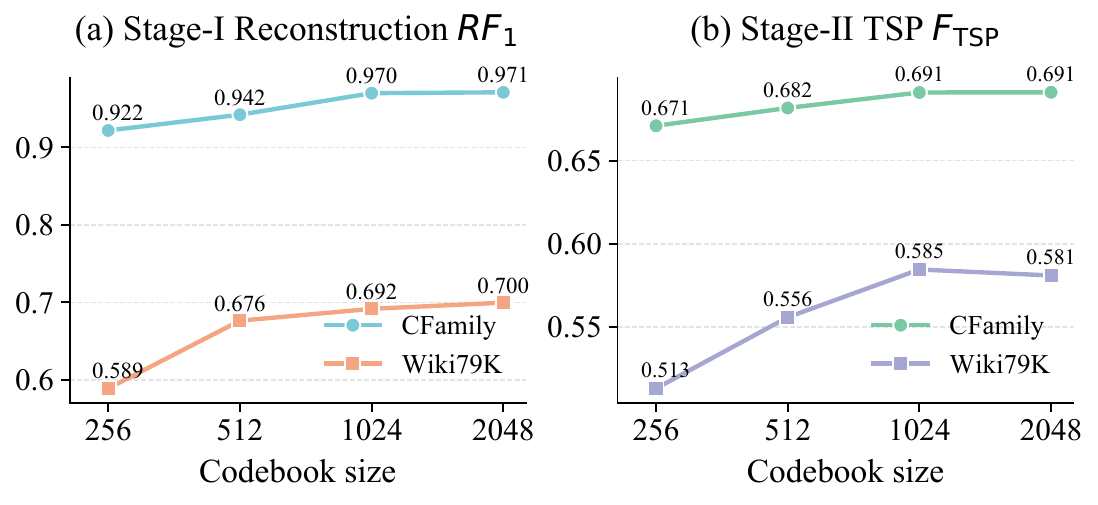}
    \caption{Effect of codebook size in CFamily and Wiki79k.}
    % \vspace{-3mm}
    \label{fig:codebook_size}
\end{figure}
\subsection{Hyperparameter Sensitivity Analysis}
\subsubsection{Effect of Codebook Size}
We study the effect of codebook size in the CRVQ module. 
As shown in Figure~\ref{fig:codebook_size}, increasing the codebook size consistently improves the reconstruction quality in Stage I.
However, the TSP performance saturates when the codebook size reaches 1024.

\subsubsection{Effect of CRVQ Configuration}
\begin{figure}
    \centering
    \includegraphics[width=0.9\linewidth]{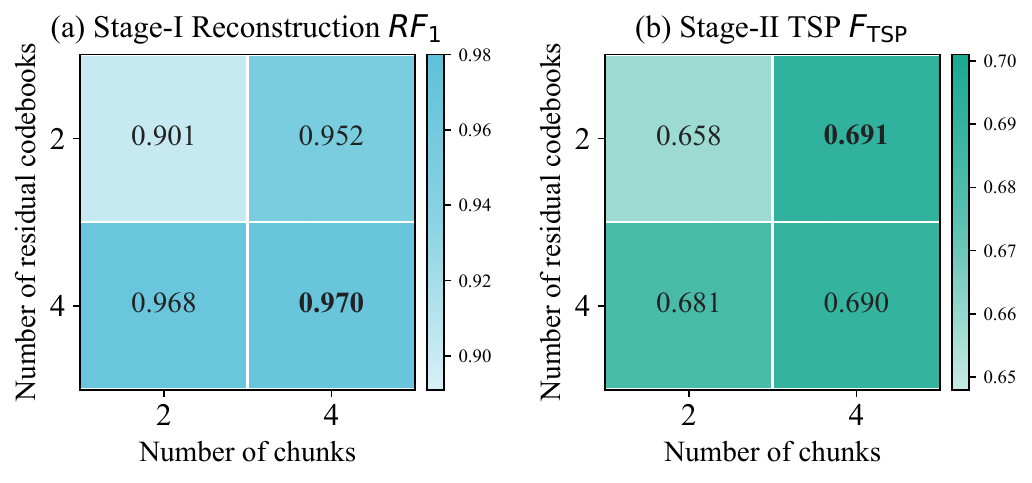}
    \caption{Performance under different CRVQ configurations in CFamily.}
    \label{fig:crvq_config}
    % \vspace{-3mm}
\end{figure}
We analyze the two key hyperparameters in CRVQ: the number of chunks and the number of residual quantization levels.  
As shown in Figure~\ref{fig:crvq_config},  increasing the number of chunks substantially improves reconstruction quality, while a moderate number of residual codebooks further benefits TSP performance. 

\subsubsection{Effect of Diffusion Steps}
\begin{figure}
    \centering
    \includegraphics[width=0.6\linewidth]{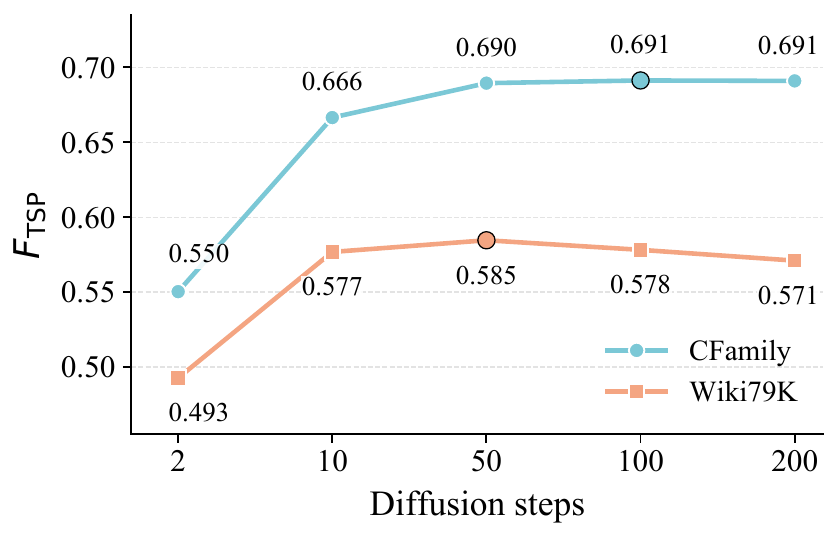}
    % \vspace{-3pt}
    \caption{Effect of diffusion steps in CFamily and Wiki79k.}
    \label{fig:diffusion_steps}
    % \vspace{-5pt}
    
\end{figure}
We further investigate the effect of the number of diffusion steps. As shown in Figure~\ref{fig:diffusion_steps}, using only 2 diffusion steps leads to clearly inferior performance. When the number of steps increases from 2 to 10 and then to 50, the $F_{\rm TSP}$ score improves substantially. However, the performance saturates after 50 or 100 steps, and further increasing the steps to 200 does not yield consistent gains. 
% , indicating that the reverse diffusion process can progressively refine the latent representation
 % and gradually recover missing triples 
 
\begin{table*}[t]
\centering
\caption{Partition-induced fragmentation statistics on three datasets.}
\label{tab:partition_stats}
\scalebox{0.8}{
\begin{tabular}{l|rrrr|rrrr}
\toprule
Dataset 
& \#Train 
& \#Retained 
& \#Uncovered 
& Uncovered Ratio
& \#Test 
& \#Same-subgraph
& \#Partition-boundary
& Boundary Ratio \\
\midrule
CFamily  
& 16,549 
& 16,514 
& 35 
& 0.21\%
& 4,598 
& 4,551 
& 47 
& 1.02\% \\
Wiki79k  
& 57,033 
& 53,937 
& 3,096 
& 5.43\%
& 15,843 
& 12,452 
& 3,391 
& 21.40\% \\
Wiki143k 
& 103,415 
& 88,984 
& 14,431 
& 13.95\%
& 28,727 
& 16,781 
& 11,946 
& 41.58\% \\

\bottomrule
\end{tabular}}
\end{table*}

\subsection{Case Study}

\begin{table}[t]
\centering
\caption{Cases of cross-subgraph triples.}
\label{tab:case_study}
\small
\resizebox{\linewidth}{!}{
\begin{tabular}{l|ccc}
\toprule
Target Triple & GPHT & DiffTSP & \modelname{} \\
\midrule
(\textit{October}, follows, \textit{September})
& Missed & Missed & Recalled $(s=0.98)$ \\

(\textit{2006 Asian Games}, follows, \textit{2002 Asian Games})
& Missed & Missed & Recalled $(s=0.97)$ \\

(\textit{Ch\^alons-en-Champagne}, twinned with, \textit{Neuss})
& Missed & Missed & Recalled $(s=0.95)$ \\

(\textit{\'{E}pinal}, located in, 
\textit{Vosges})
& Missed & Missed & Recalled $(s=0.91)$ \\

(\textit{Braga}, shares border with, \textit{Guimar\~aes})
& Missed & Missed & Recalled $(s=0.89)$ \\

(\textit{Bonn}, country, \textit{Weimar Republic})
& Missed & Missed & Recalled $(s=0.75)$ \\

\bottomrule
\end{tabular}
}
\end{table}

We provide a qualitative case study to illustrate the failure mode of subgraph-based TSP baselines. A cross-subgraph triple refers to a test triple whose head and tail entities do not appear in any common subgraph of subgraph-based methods. 
Table~\ref{tab:case_study} shows representative cross-subgraph test triples recalled by \modelname{} using the same validation-selected threshold as in the main evaluation. 
The compared subgraph-based methods do not generate these triples.

\subsection{Analysis of Cross Subgraph Recovery}
A key motivation of our framework is that subgraph partitioning may fragment relational dependencies. 
To directly examine this issue, we construct a cross-subgraph test subset based on the partitions used by the subgraph-based baseline. 
A test triple $(h,r,t)$ is considered cross-subgraph if its head entity $h$ and tail entity $t$ are not contained in any common subgraph. 
Such triples are difficult for subgraph-based methods because the evidence associated with $h$ and $t$ is separated by the partition boundary.

We examine how much structural information is affected by the partition process. 
Given the subgraph collection used by the subgraph-based baseline, we define a training triple as covered if it appears in at least one subgraph. Otherwise, this triple is counted as partition-uncovered. Similarly, a test triple is defined as cross-subgraph if its head and tail entities do not appear in any common subgraph.
Table~\ref{tab:partition_stats} reports the statistics. 
On Wiki79k, $5.43\%$ training triples are uncovered after partitioning, and $21.40\%$ of test triples are partition-boundary triples. 
The effect is more pronounced on Wiki143k. 
% , where $13.95\%$ of training triples are uncovered and $41.58\%$ of test triples cross partition boundaries
CFamily shows smaller fragmentation, with $1.02\%$ partition-boundary test triples. 
These results suggest that the impact of subgraph partitioning varies across datasets, but can be non-negligible on larger KGs. 
This motivates our unified graph latent formulation, where all head-relation latent embeddings are represented in a globally shared code latent space.

\paragraph{Performance on partition-boundary triples.}

We further conduct a subset-level evaluation on partition-boundary test triples, whose head and tail entities do not occur in any common subgraph constructed by the subgraph-based models. As shown in Table~\ref{tab:boundary_recall}, GPHT and DiffTSP recover none of these triples. In contrast, \modelname{} can score the partition-boundary triples, resulting in ${STRecall}_{\mathrm{bd}}$ values of 0.229 and 0.195 on Wiki79k and Wiki143k, respectively. 

Partition-boundary triples remain difficult because their supporting evidence can be sparse or structurally distant, and the decoding threshold is selected according to overall validation $F_{\rm{TSP}}$ rather than boundary-specific performance. Nevertheless, the non-zero recovery of \modelname{}, compared with zero recovery by the subgraph-based methods, provides quantitative evidence that the unified latent generation framework enables the recovery of triples excluded by subgraph-restricted candidate construction. 
\begin{table}[t]
\centering
\caption{
Diagnostic evaluation on partition-boundary test triples. ${STRecall}_{\mathrm{bd}}$ is computed only over the partition-boundary subset. }
\label{tab:boundary_recall}
\resizebox{0.7\linewidth}{!}{
\begin{tabular}{llc>{\columncolor{gray!10}}c}
\toprule
Dataset
& Method
& Recalled
& ${STRecall}_{\mathrm{bd}}$ \\
\midrule
\multirow{3}{*}{Wiki79k}
& GPHT
& $0/3{,}391$ 
& 0.000 \\
& DiffTSP
& $0/3{,}391$
& 0.000 \\
& \modelname{}
& $178/3{,}391$
& \textbf{0.229} \\
\midrule
\multirow{3}{*}{Wiki143k}
& GPHT
& $0/11{,}946$
& 0.000 \\
& DiffTSP
& $0/11{,}946$
& 0.000 \\
& \modelname{}
& $456/11{,}946$
& \textbf{0.195} \\
\bottomrule
\end{tabular}
}
\end{table}

\subsection{Efficiency Study}
We further compare \modelname~ with DiffTSP and GPHT under the same hardware environment. 
As shown in Table~\ref{tab:runtime_speedup}, \modelname{} achieves faster inference.  These results show that performing generation in the latent space can provide favorable inference efficiency despite the iterative nature of diffusion sampling.
\modelname{} also avoids explicitly constructing the raw
$|\mathcal{E}|\times|\mathcal{R}|\times|\mathcal{E}|$ adjacency tensor and instead performs generation over
$\mathbf{Z}\in\mathbb{R}^{|\mathcal{E}|\times|\mathcal{R}|\times d}$. Moreover, head-wise chunking reduces peak latent memory without partitioning the graph or introducing separate local model parameters.  More details are in the Appendix.
\paragraph{More Experimental Details.} Additional analysis on ablation studies, implementation details, observed ratios, and inference memory consumption are reported in the Appendix.
\begin{table}[t]
\centering
\caption{
Inference time (s) comparison and speedup.
% All methods are evaluated on a single A100 GPU.
% \modelname{} uses $T=50$ diffusion steps.
The speedup is computed against the fastest baseline DiffTSP.
}
\label{tab:runtime_speedup}
\resizebox{0.8\linewidth}{!}{
\begin{tabular}{lrrrr}
\toprule
Dataset 
& GPHT 
& DiffTSP 
& \modelname{} 
& Speedup 
\\
\midrule
CFamily  
& 46.25  
& 8.13   
& 1.64  
& 4.9$\times$ 
\\
Wiki79k  
& 151.94 
& 83.77  
& 12.79 
& 6.5$\times$ 
\\
Wiki143k 
& 583.69 
& 101.07 
& 35.93 
& 2.8$\times$ 
\\
\bottomrule
\end{tabular}
}
\end{table}

% \textbf{Codebook utilization.}
% We report codebook perplexity and usage ratio to analyze whether the residual codebooks are effectively used.

\section{Conclusion}
In this work, we proposed \modelname, a two-stage conditional latent generative framework, for knowledge graph triple set prediction. \modelname~first learns a relation-aware residual-quantized latent space, and then performs observation-anchored diffusion in this space to recover missing relational residuals. Instead of relying on subgraph-level local prediction, \modelname~performs completion in a globally shared latent space. 
Experimental results demonstrate that \modelname~achieves state-of-the-art performance. This work focuses on establishing the effectiveness of unified latent generation for TSP, and a systematic investigation of its scalability to larger knowledge graphs is left for future work.

\clearpage

\bibliographystyle{ACM-Reference-Format}
\bibliography{bio}

\appendix
\input{appendixd}

\clearpage
\end{document}

%% file: appendixd.tex
\section{Method Details}
\subsection{Dual-view Linear Attention}
Given a sequence $\mathbf{X}\in\mathbb{R}^{L\times d}$, each attention head computes
\begin{equation}
    \mathbf{Q}=\mathbf{X}\mathbf{W}_Q,\quad
\mathbf{K}=\mathbf{X}\mathbf{W}_K,\quad
\mathbf{V}=\mathbf{X}\mathbf{W}_V,
\end{equation}
and uses the positive feature map $\phi(x)=\operatorname{ELU}(x)+1$:
\begin{equation}
\operatorname{LinAttn}(\mathbf{X})_i
=
\frac{
\phi(\mathbf{q}_i)^\top
\left(\sum_{j=1}^{L}\phi(\mathbf{k}_j)\mathbf{v}_j^\top\right)
}{
\phi(\mathbf{q}_i)^\top
\left(\sum_{j=1}^{L}\phi(\mathbf{k}_j)\right)+\epsilon
}.
\end{equation}
We employ multi-head attention and concatenate the outputs of all heads. Relation-wise attention applies this operation over the $R$ relation tokens for each entity, whereas node-wise attention applies it over the $N$ entity tokens for each relation. The two branches use independent projection parameters.
\subsection{Complexity Analysis}
\label{sec:complexity}

Let $N=|\mathcal{E}|$, $R=|\mathcal{R}|$, and $d$ denote the entity count, relation count, and latent dimension. Let $L_e$, $L_d$, and $L_c$ denote the encoder, denoiser, and residual quantization depths, $C$ the codebook size, and $S$ the number of reverse diffusion steps.
Raw adjacency space reconstruction or diffusion requires $O(N^2R)$ memory and at least $O(N^2Rd)$ computation. Instead, \modelname{} generates head-relation latents $\mathbf{Z}\in\mathbb{R}^{N\times R\times d}$ with $O(NRd)$ latent memory.

\noindent\textbf{Stage I.}
For input triples $\mathcal{T}_{\rho}$, sparse projection costs $O(|\mathcal{T}_{\rho}|d)$, while reconstruction with $K_{\mathrm{neg}}$ negatives per positive costs $O((1+K_{\mathrm{neg}})|\mathcal{T}_{\rho}|d)$.
Relation-wise encoding and CRVQ respectively cost
\begin{equation}
O\!\left(L_eN(Rd^2+R^2d)\right)
\quad\text{and}\quad
O(NRL_cCd).
\end{equation}
CRVQ uses $O(L_cCd)$ codebook memory. Splitting embeddings into proportionally smaller chunks leaves its leading order computation and parameter memory unchanged.

\noindent\textbf{Stage II.}
Dual-view linear attention avoids dense $R\times R$ and $N\times N$ attention maps and their $O(NR^2d+RN^2d)$ computation.
With attention feature dimension and FFN width both $O(d)$, each denoiser evaluation costs $O(L_dNRd^2)$.
Training samples one timestep and has the same per-instance asymptotic cost, whereas $S$-step DDPM sampling costs $O(SL_dNRd^2).$
Reverse steps execute sequentially without accumulating activations across steps, so peak memory does not scale with $S$.

\noindent\textbf{Chunking and decoding.}
Scoring a candidate set $\mathcal{C}$ costs $O(|\mathcal{C}|d)$, or $O(N^2Rd)$ for exhaustive decoding, while scores can be generated and thresholded incrementally.
In both stages, processing $B_h$ head entities per chunk reduces active latent memory and exhaustive score buffer memory, respectively, as
\begin{equation}
\begin{aligned}
O(NRd) \rightarrow O(B_hRd), \quad
O(N^2R) \rightarrow O(B_hRN).
\end{aligned}
\end{equation}
Total asymptotic computation remains unchanged.
Chunking manages memory rather than partitioning the graph: all chunks share the encoder, codebooks, denoiser, and decoder within a common latent space.
Global node-wise interactions can be preserved by accumulating key-value statistics across chunks rather than computing attention independently within each chunk.

\subsection{Stage-I Reconstruction Fidelity Metrics}
Stage I learns a compact latent representation by reconstructing its input graph. Its reconstruction metrics measure how faithfully an input is encoded and decoded. The diagnostic evaluation below supplies an evaluation graph containing test triples to the encoder, so its scores describe reconstruction with access to those triples. These diagnostic inputs and scores are excluded from TSP training and hyperparameter selection.
Stage-I reconstruction evaluation is performed by feeding an evaluation KG containing both training and test triples:
\begin{equation}
    \mathcal{G}_{\mathrm{eval}}
    =
    (\mathcal{E},\mathcal{R},\mathcal{T}_{\mathrm{eval}}),
    \quad
    \mathcal{T}_{\mathrm{eval}}
    =
    \mathcal{T}_{\mathrm{train}}
    \cup
    \mathcal{T}_{\mathrm{test}}.
\end{equation}
After encoding, quantization, and decoding, we obtain a reconstructed triple set 
$\mathcal{T}_{\mathrm{rec}}$. 
Since training triples are already trained in our model and should not inflate the reconstruction score, we remove them from the reconstructed set:
\begin{equation}
    \mathcal{T}^{\mathrm{eval}}_{\mathrm{rec}}
    =
    \mathcal{T}_{\mathrm{rec}}
    \setminus
    \mathcal{T}_{\mathrm{train}}.
\end{equation}
The reconstruction metrics are then computed only against the test triples:
\begin{equation}
    {RecPrecision}
    =
    \frac{
    |\mathcal{T}^{\mathrm{eval}}_{\mathrm{rec}}\cap \mathcal{T}_{\mathrm{test}}|
    }
    {
    |\mathcal{T}^{\mathrm{eval}}_{\mathrm{rec}}|
    },
\end{equation}
\begin{equation}
    {RecRecall}
    =
    \frac{
    |\mathcal{T}^{\mathrm{eval}}_{\mathrm{rec}}\cap \mathcal{T}_{\mathrm{test}}|
    }
    {
    |\mathcal{T}_{\mathrm{test}}|
    },
\end{equation}
\begin{equation}
    RF_1
    =
    \frac{
    2\cdot {RecPrecision}\cdot {RecRecall}
    }
    {
    {RecPrecision}+{RecRecall}
    }.
\end{equation}
Here, $RF_1$ is used as the primary diagnostic metric for Stage-I reconstruction analysis.

\section{Experiment Details}
\subsection{Baselines}
We select both link prediction methods and TSP methods as baselines. The representative link prediction methods, including  HAKE~\cite{zhang2020hake}, PairRE~\cite{chao2020pairre}, RGCN~\cite{2018rgcn}, CompGCN~\cite{CompGCN}, AstarNet~\cite{zhu2023astarnet} and ULTRA~\cite{2024ultra}, are modified for TSP by scoring all candidate triples in the graph and selecting those above a threshold as the predicted results. The threshold is determined through grid search on the validation set to achieve the best performance. 
The latest TSP methods, such as GPHT~\cite{zhang2024gpht}, LLMTSP~\cite{yuan2024llmtsp}, SMuRT~\cite{yuan2026smurttsp} and DiffTSP~\cite{guan2026difftsp}, are also included.

\subsection{Evaluation Metrics}
\paragraph{TSP Evaluation Metrics.}
$\textit{JPrecision}$ measures the precision of the predicted triple set while accounting for the size of the set itself. 
$\textit{STRecall}$ measures the percentage of test triples successfully recovered by the model with a square-root transformation. 
$F_{\rm{TSP}}$ is the harmonic mean of $\textit{JPrecision}$ and $\textit{STRecall}$.
Higher values indicate better performance for all metrics. 

Following GPHT, we denote the predicted missing triple set as $\mathcal{T}_{\mathrm{pred}}$. 
This corresponds to $\widehat{\mathcal{T}}_{\mathrm{miss}}$ in our problem definition. 
Under the closed-world assumption (CWA), the predicted triples are divided into correct and incorrect parts:

\begin{small}
\begin{align}
    \mathcal{T}^{\mathrm{CWA}+}_{\mathrm{pred}} 
    &= 
    \mathcal{T}_{\mathrm{pred}} \cap \mathcal{T}_{\mathrm{test}}, \\
    \mathcal{T}^{\mathrm{CWA}-}_{\mathrm{pred}} 
    &= 
    \mathcal{T}_{\mathrm{pred}} \setminus 
    \mathcal{T}^{\mathrm{CWA}+}_{\mathrm{pred}}, \\
    \mathcal{T}^{\mathrm{CWA}}_{\mathrm{pred}} 
    &= 
    \mathcal{T}^{\mathrm{CWA}+}_{\mathrm{pred}} 
    \cup 
    \mathcal{T}^{\mathrm{CWA}-}_{\mathrm{pred}}
    =
    \mathcal{T}_{\mathrm{pred}} .
\end{align}
\end{small}

Under the relation-similarity partial open-world assumption (RS-POWA), the positive predictions are defined as

\begin{small}
\begin{align}
    \mathcal{T}^{\mathrm{POWA}+}_{\mathrm{pred}}
    &=
    \mathcal{T}_{\mathrm{test}} \cap \mathcal{T}_{\mathrm{pred}} .
\end{align}
\end{small}
The negative part is defined as
\begin{small}
\begin{align}
    \mathcal{T}^{\mathrm{POWA}-}_{\mathrm{pred}}
   & =
    \bigl\{
    (h,r,t) \mid 
    (h,r,t)\in \mathcal{T}_{\mathrm{pred}},\ 
    (h,r,t)\notin \mathcal{T}_{\mathrm{test}}, \nonumber \\
    &\exists r'\in\mathcal{R},\ 
    (h,r',t)\in
    (\mathcal{T}_{\mathrm{train}}\cup\mathcal{T}_{\mathrm{test}})
    \ \wedge\
    \operatorname{sim}(r,r')<\theta
    \bigr\}.
\end{align}
\end{small}
Then the valid predicted set under RS-POWA is

\begin{small}
\begin{align}
    \mathcal{T}^{\mathrm{POWA}}_{\mathrm{pred}}
    =
    \mathcal{T}^{\mathrm{POWA}+}_{\mathrm{pred}}
    \cup
    \mathcal{T}^{\mathrm{POWA}-}_{\mathrm{pred}} .
\end{align}
\end{small}

For $WA\in\{\mathrm{CWA},\mathrm{POWA}\}$, the three metrics are computed as
\begin{small}
\begin{align}
    \textit{JPrecision}
    &=
    \frac{1}{2}
    \left(
    \frac{
    |\mathcal{T}^{WA+}_{\mathrm{pred}}|
    }{
    |\mathcal{T}^{WA}_{\mathrm{pred}}|
    }
    +
    \frac{
    |\mathcal{T}^{WA+}_{\mathrm{pred}}|
    }{
    |\mathcal{T}_{\mathrm{pred}}|
    }
    \right), \\
    \textit{STRecall}
    &=
    \left(
    \frac{
    |\mathcal{T}^{WA+}_{\mathrm{pred}}|
    }{
    |\mathcal{T}_{\mathrm{test}}|
    }
    \right)^{\frac{1}{2}}, \\
    F_{\rm{TSP}}
    &=
    \frac{
    2\cdot \textit{STRecall}\cdot \textit{JPrecision}
    }{
    \textit{STRecall}+\textit{JPrecision}
    } .
\end{align}
\end{small}

Since $F_{\rm{TSP}}$ is the harmonic mean of $JPrecision$ and $STRecall$, it summarizes the trade-off between set precision and test triple recovery. 
Therefore, we use $F_{\rm{TSP}}$ as the primary metric for model comparison, following the TSP evaluation protocol in GPHT. 
$JPrecision$ and $STRecall$ are reported as complementary metrics to analyze whether the improvement comes from more accurate predicted sets or better recovery of test triples. 
Unless otherwise specified, our discussion mainly focuses on $F_{\rm{TSP}}$.
We report micro-averaged results over all predicted triples. 
% For ranking-based baselines, we select the same number of predicted triples as the target set or use a validation-tuned score threshold.

\subsection{Triple Set Decoding and Threshold Selection}

For each method, the final predicted triple set is obtained by thresholding its predicted triple scores. 
Let $s_m(h,r,t)$ denote the score assigned by method $m$ to a candidate triple $(h,r,t)$. 
Given a threshold $\tau$, the predicted triple set is
\begin{equation}
    \mathcal{T}_{\mathrm{pred}}(\tau)
    =
    \{(h,r,t)\in \mathcal{C} \mid s_m(h,r,t) \ge \tau\},
\end{equation}
where $\mathcal{C}$ denotes the candidate triple set after filtering out observed training triples.
Following the evaluation protocol used in GPHT and DiffTSP, we select the threshold on the validation set for each method and each dataset. 
Specifically, we search over a predefined threshold grid $\Theta$ and choose
\begin{equation}
    \tau^{*}_{m}
    =
    \arg\max_{\tau\in\Theta}
    F_{\mathrm{TSP}}^{\mathrm{val}}
    \bigl(
    \mathcal{T}_{\mathrm{pred}}(\tau)
    \bigr).
\end{equation}
The selected threshold $\tau^{*}_{m}$ is then fixed and used to generate the predicted triple set on the test set. 
All reported ${JPrecision}$, ${STRecall}$, and $F_{\mathrm{TSP}}$ values are computed on the test set using this validation-selected threshold. 
This protocol is applied consistently to \modelname{} and all compared baselines to ensure a fair comparison.

\subsection{Implementation Details}
All experiments are conducted on one NVIDIA A100 GPU. We use the Adam optimizer. Hyperparameters are selected by grid search on the validation set, more details are in Table~\ref{tab:hyperdetail}.
To further reduce peak memory usage, both Stage I and Stage II support head-wise chunking implementation. 
Specifically, the latent tensor is processed in chunks along the head-entity dimension instead of materializing all head-relation embeddings at once. 
This is an engineering strategy for memory-efficient training and inference, not a subgraph partitioning strategy. 
% All chunks share the same encoder, quantizer codebooks, diffusion denoiser, and triple decoder. 
% Therefore, the latent variables remain in a globally shared code space, and triple scores are decoded under the same global parameters. 
% This differs from subgraph-based TSP methods, where each subgraph may be predicted locally.
% we use the Adam optimizer
\begin{table}[t]
\centering
\caption{Training hyperparameters on three datasets.}
\label{tab:hyperdetail}
\resizebox{0.9\linewidth}{!}{
\begin{tabular}{lccc}
\toprule
Hyperparameter & CFamily & Wiki79k & Wiki143k \\
\midrule
\multicolumn{4}{c}{Stage-I Training} \\
\midrule
Encoder hidden dimension & 64 & 64 & 32 \\
Chunks of CRVQ $K$ & 4 & 4 & 4 \\
Number of residual quantizers $L_c$ & 4 & 4 & 2 \\
Latent dimension & 16 & 16 & 16 \\
Codebook size $C$ & 1024 & 1024 & 2048 \\
Commitment loss weight $\lambda_m$ & 1 & 1 & 1 \\
CRVQ loss weight ${\lambda}_{\mathrm{VQ}}$ & 0.25 & 0.25 & 0.25 \\
Number of negative samples $K_{\rm neg}$ & 512 & 256 & 256 \\
Learning rate & $5\times 10^{-4}$ & $5\times 10^{-4}$ & $5\times 10^{-4}$  \\
\midrule
\multicolumn{4}{c}{Stage-II Training} \\
\midrule
Ratio $\rho_{\rm min}$& 0.7 & 0.7 & 0.7 \\
Observed KG ratio $\rho_{\rm obs}$& 0.8 & 0.8 & 0.8 \\
Variance schedule & Linear & Linear & Linear \\
Variance schedule $\beta_{\min}$
& $1\times10^{-3}$
& $2\times10^{-3}$
& $2\times10^{-3}$ \\
Variance schedule $\beta_{\max}$
& $2\times10^{-1}$
& $4\times10^{-1}$
& $4\times10^{-1}$ \\
Diffusion steps & 100 & 50 & 50 \\
Denoising hidden dimension & 128 & 128 & 64 \\
Condition dimension & 128 & 128 & 64 \\
Time embedding dimension & 128 & 128 & 64 \\
Denoiser depth $L$ & 3 & 2 & 2 \\
Learning rate & $5\times 10^{-4}$ & $5\times 10^{-4}$ & $5\times 10^{-4}$  \\
\bottomrule
\end{tabular}
}
\end{table}

\subsection{Ablation Study Details}
Table~\ref{tab:stage1_ablation} reports the Stage-I reconstruction ablations.
For the VQ-VAE comparison, we match the total learnable codebook parameter budget while keeping the encoder, decoder, latent dimensionality, and training protocol identical to those of \modelname{}.
Replacing CRVQ with VQ-VAE causes the largest drop in reconstruction fidelity, supporting the effectiveness of CRVQ in preserving KG structure.
Removing relational-view attention or replacing the factorized decoder also reduces reconstruction quality, highlighting the contributions of relation-aware encoding and factorized decoding.

Although the Vanilla AE achieves slightly higher reconstruction fidelity, its TSP performance is substantially worse (Table~\ref{tab:ablation}).
This contrast suggests that reconstruction fidelity alone is insufficient for effective latent generation and is consistent with a regularizing role of quantization.
Stage-I reconstruction fidelity is a diagnostic of latent representation quality rather than a strict upper bound on TSP performance, since Stage II directly decodes continuous latent predictions without re-quantization.

\begin{table}[t]
\centering
\caption{
Stage-I reconstruction ablation results.
We report reconstruction $RF_1$ on three datasets.
}
\label{tab:stage1_ablation}
\resizebox{1\linewidth}{!}{
\begin{tabular}{lccc}
\toprule
Variant 
& CFamily 
& Wiki79k 
& Wiki143k \\
\midrule
\modelname{} 
& \text{0.9703} 
& \text{0.6918} 
& \text{0.7145} \\
w/o Relational view Attention 
& 0.9057 
& 0.6229 
& 0.6927 \\
w/o CRVQ (VQ-VAE)
& 0.4591 
& 0.0575 
& 0.1098 \\
w/o Quantization (Vanilla AE)
& 0.9898
& 0.7187 
& 0.7339 \\
w/o Factorized Decoder 
& 0.8895 
& 0.5891 
& 0.6219 \\
\bottomrule
\end{tabular}
}
\end{table}

\subsection{Sensitivity to the Observed Ratio}
\begin{figure}
    \centering
    \includegraphics[width=0.8\linewidth]{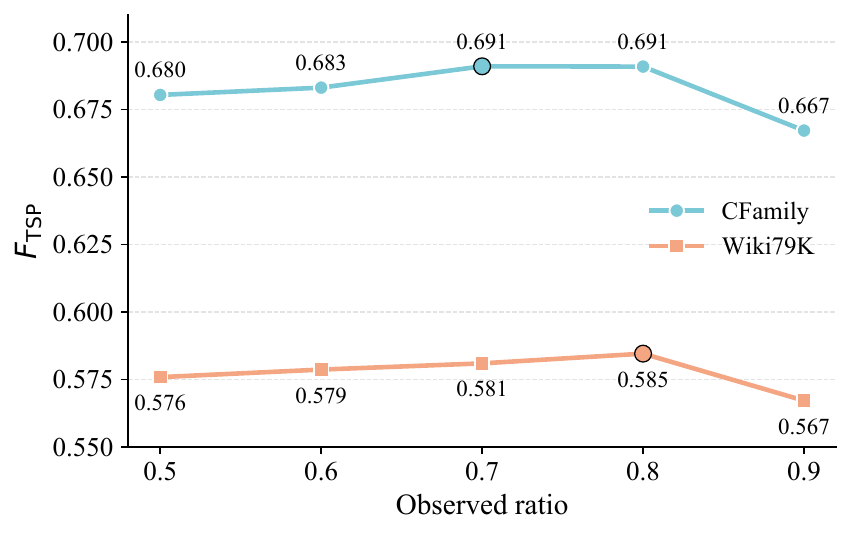}
    \caption{Sensitivity to the observed ratio $\rho_{\rm obs}$.
We vary the ratio of observed triples from 0.5 to 0.9 and report $F_{\mathrm{TSP}}$ on CFamily and Wiki79k.
}
    \label{fig:obs_ratio}
\end{figure}

We further study the effect of the observed ratio, i.e., the proportion of triples retained as the observed graph for conditioning in the diffusion training process. Figure~\ref{fig:obs_ratio} shows that the performance is relatively stable across different observed ratios, but the best results are achieved when the observed ratio is around 0.7--0.8. When the observed ratio becomes too high, the prediction task becomes less informative because fewer target triples remain to be recovered, and the model may also receive reduced supervision on the missing part. These results suggest that a moderate observed ratio provides a good balance between observation information and prediction difficulty. Therefore, we use an observed ratio of 0.8 as the default setting.

\subsection{Effect of CRVQ Configurations}
\begin{figure}
    \centering
    \includegraphics[width=1\linewidth]{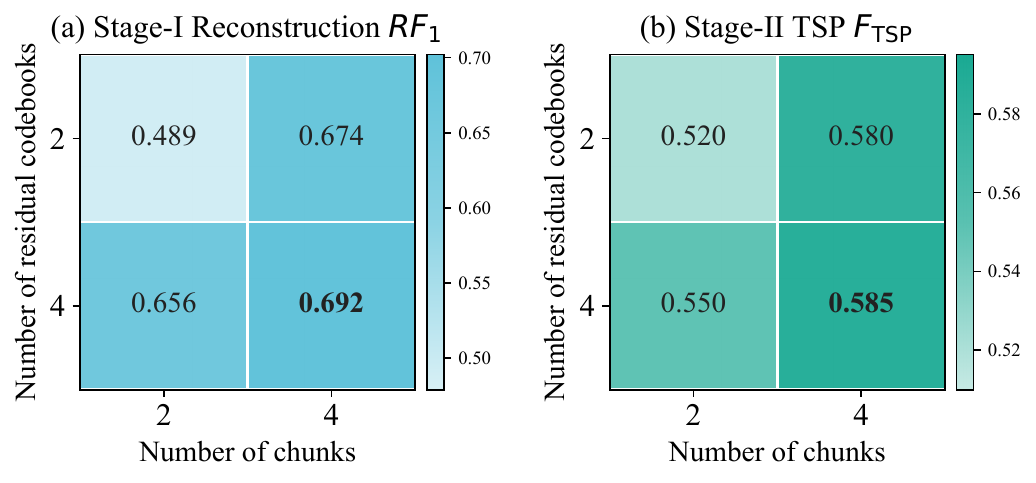}
    \caption{Performance under different CRVQ configurations in Wiki79k.
}
    \label{fig:hypercrvqwiki}
\end{figure}

As shown in Figure~\ref{fig:hypercrvqwiki}, we also analyze two key hyperparameters of CRVQ on Wiki79k. Increasing either the number of chunks or the number of residual codebooks improves both Stage-I reconstruction fidelity and TSP performance.

\subsection{Inference Efficiency and Memory Cost}
\begin{table}[t]
\centering
\caption{
Inference efficiency of \modelname{} under different diffusion steps. 
% Prediction time is measured in seconds, and peak memory is measured in MB on a single A100 GPU.
}
\label{tab:lkgdiff_efficiency}
\resizebox{0.9\linewidth}{!}{
\begin{tabular}{lrrrr}
\toprule
Dataset 
& \multicolumn{2}{c}{Prediction Time (s)}
& \multicolumn{2}{c}{Peak Memory (MB)} \\
\cmidrule(lr){2-3}\cmidrule(lr){4-5}
& $S=50$ & $S=100$ & $S=50$ & $S=100$ \\
\midrule
CFamily  & 0.72  & 1.64  & 2,136  & 2,136 \\
Wiki79k  & 12.79 & 22.96 & 12,366 & 12,366 \\
Wiki143k & 35.93 & 45.76 & 19,872 & 19,872 \\
\bottomrule
\end{tabular}
}
\end{table}

We evaluate the inference efficiency of \modelname~ on one A100 GPU. Table~\ref{tab:lkgdiff_efficiency} reports the prediction time and peak memory under different diffusion steps. 
The inference time increases with $S$, since DDPM sampling requires iterative reverse denoising. 
In contrast, the peak memory remains unchanged when increasing $S$, because the reverse process is executed sequentially and does not store activations from all denoising steps.